%% file: main.tex
\pdfoutput=1

\documentclass[10pt,twocolumn,letterpaper]{article}

\usepackage[pagenumbers]{cvpr}      %

\input{preamble}

\definecolor{cvprblue}{rgb}{0.21,0.49,0.74}
\usepackage[pagebackref,breaklinks,colorlinks,allcolors=cvprblue]{hyperref}

\def\confName{CVPR}
\def\confYear{2027}

\title{Tripwire: Robot Agents that Program Their Own Interrupts\\under Human Disturbance}

\author{Sunli Chen$^{1*}$ \quad Yun Zong$^{2*}$ \quad Qinhong Zhou$^{1}$ \quad Xiaowen Qiu$^{1}$\\ Jiaben Chen$^{1}$  \quad Zeyuan Yang$^{1}$ \quad Chuang Gan$^{1}$\\
$^{1}$University of Massachusetts at Amherst \quad $^{2}$University of Minnesota - Twin Cities\\[2pt]
{\small $^*$Equal contribution}
}

\begin{document}
\twocolumn[{%
\renewcommand\twocolumn[1][]{#1}%
\maketitle
\begin{center}
  \centering
  \captionsetup{type=figure}
  \includegraphics[width=\textwidth]{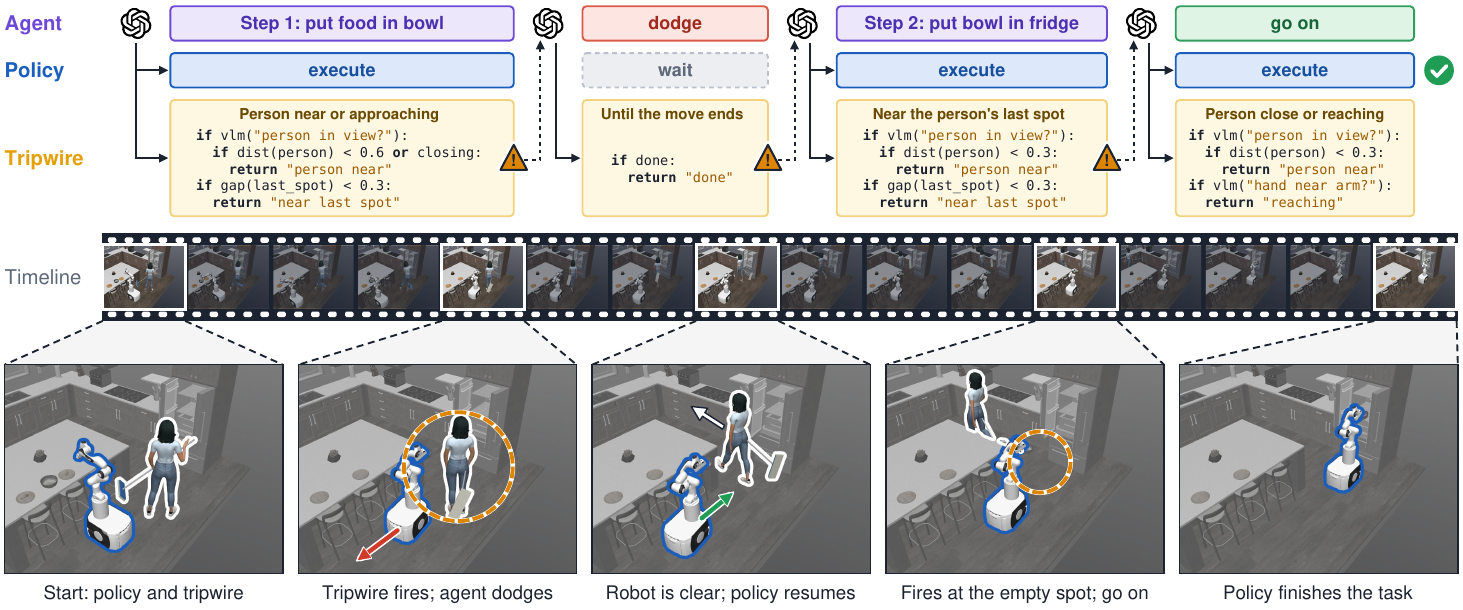}
  \captionof{figure}{TRIP in one episode with Xiaomi-Robotics-1, while a person mops past the robot. A tripwire fires when the person comes within 0.3\,m of the base, and the agent backs the robot out of the way, then resumes the policy under a tripwire on the person's new position. When that one fires at the spot the person has just left, the agent rewrites it to watch for a close or reaching person, and the policy finishes the task. Top: when the agent, the policy and each (simplified) tripwire run.}
  \label{fig:teaser}
\end{center}%
\vspace{1em}%
}]
\input{sec/0_abstract}    
\input{sec/1_intro}
\input{sec/2_related}

\input{sec/3_method}

\input{sec/3_dataset}
\input{sec/4_experiments}

\input{sec/7_conclusion}

{
    \small
    \bibliographystyle{ieeenat_fullname}
    \bibliography{main}
}

\input{sec/X_suppl}

\end{document}

%% file: preamble.tex
\makeatletter
\newcommand{\csl}[1]{{\color{red}[\def\csl@sep{}\@for\csl@key:=#1\do{\csl@sep ??\def\csl@sep{, }}]}}
\makeatother

\ifdefined\pdfsuppressptexinfo\pdfsuppressptexinfo=-1\fi

\usepackage{algorithm}
\usepackage[noend]{algpseudocode}
\usepackage{multirow}
\usepackage{colortbl}
\definecolor{topblue}{RGB}{221,235,250}

%% file: sec/0_abstract.tex
\begin{abstract}
Agent harnesses are turn-based: the agent acts, waits for the result, then reasons. When the action invokes a pretrained robot policy, such as a vision-language-action (VLA) or world action model (WAM), the policy may react continuously to observations, but the agent cannot revisit its decision until the call returns. Meanwhile, a person may step into the robot's path or move the object it is about to grasp. Invoking the agent more often narrows this blind interval, but a reasoning model takes seconds per decision and most control steps need no new one. We propose TRIP, a training-free, event-driven harness in which the agent programs its own interrupts. With each call, the agent writes a tripwire: a lightweight, stateful Python program that runs alongside the policy, can query a compact VLM (for example, to locate a person and measure their distance), and decides whether the call continues or control returns to the agent. When it fires, the harness halts the policy and returns control with the reason; the agent then reassesses, adapts, and continues under a rewritten tripwire.
We evaluate TRIP with three VLAs and one WAM on RoboHD, a benchmark of 168 RoboCasa365 scenes in which a simulated person disturbs the robot and any human-robot contact counts as failure. Compared with a turn-based harness using the same agent and tools, TRIP halves the rate of human contact with the strongest policy, raising safety from 61.3\% to 81.3\%, without sacrificing task success (38.7\% vs.\ 31.0\%). Real-robot experiments show that TRIP completes two manipulation tasks under human interference where the turn-based harness and the policy alone largely fail.
\end{abstract}

%% file: sec/1_intro.tex
\section{Introduction}
\label{sec:intro}
Frontier vision-language models increasingly act as agents for robots: they interpret scenes, decompose tasks, and coordinate execution through tool calls~\cite{team2026systematically, yang2026astranomy, chen2026showharness, wang2026auriga}. A common harness exposes a pretrained manipulation policy as a callable skill and invokes the agent again only when the call returns~\cite{zhang2026harnessvla}. This turn-based interface locks the agent out while the policy runs, even though an action chosen at the start of a call can become unsafe or obsolete before it finishes. If a person steps into the robot's path mid-execution, the agent remains blind to the disturbance until control is handed back.

The policy's own closed loop does not remove this blind spot. Vision-language-action (VLA) models~\cite{black2024pi_0, intelligence2025pi_, bjorck2025gr00t} and world action models (WAMs)~\cite{ye2026world} can react to new observations, but strong performance when nothing interferes with the task does not guarantee reliable behavior around people. In a shared kitchen, someone may move the target plate or close the fridge door the robot has just opened. When we augment RoboCasa365 tasks~\cite{nasiriany2026robocasa365} with a reactive human avatar, all four pretrained policies we evaluate lose more than half of their successes; the strongest drops from 61.9\% to 20.8\% and touches the person in roughly half of all episodes. The agent can interpret such disturbances and choose a response, but it needs timely access to the changing scene. Querying it more frequently incurs substantial computation, and each high-level decision still requires an inference call that can take seconds~\cite{team2026systematically, yang2026astranomy}.

Watching the scene does not require strong reasoning: noticing that a person has come close is a perception problem that a compact VLM and a depth image settle in a fraction of a second. Existing systems watch a running call in one of three ways. VoLo~\cite{chen2026volo} polls a frontier model every few seconds. DoReMi~\cite{guo2024doremi} asks a small VLM whether language constraints written by the planner still hold: it can look for anything the planner names, but it gets back only yes or no. Code-as-Monitor~\cite{zhou2025code} runs generated code over geometric states tracked from the start of a subgoal, without issuing new VLM queries during execution. Reacting to a person takes both: spotting someone who was not in the scene when the check was written, and judging from their distance and motion whether they matter. What should interrupt the agent also shifts with its latest decision. After judging that a nearby person does not obstruct the task, the agent resumes the policy and needs to hear only if the person comes closer; if it chooses to wait, it instead needs to know when the person leaves. We let the agent program the interrupt and update it as its decisions change.

We introduce \textbf{TRIP} (\cref{fig:teaser}), a harness in which robot agents program their own interrupts. The agent attaches a \emph{tripwire} to a tool call: a stateful Python program that runs alongside policy execution, direct robot motions, or waiting. A tripwire both looks and computes: it chooses which cameras to read and what to ask a compact VLM, such as where a person is, and combines the answer with depth, memory of earlier checks, and timers to decide whether to interrupt. When it fires, the harness halts the active call before waking the agent with the reason, so stopping costs one check (0.24\,s on average on our real robot) and not an agent turn (4.3\,s). The agent then resumes the policy, repositions the robot, or waits, updating the tripwire as needed so it wakes the agent again when the situation changes. Because the policy remains a black box, TRIP wraps pretrained VLAs and WAMs without fine-tuning.

We evaluate TRIP in simulation with three VLAs and one WAM on \textbf{RoboHD}, a benchmark of 168 scenes built by adding a reactive human avatar to 42 RoboCasa365 tasks~\cite{nasiriany2026robocasa365}. RoboHD covers spatial, object-state, task-state, and perceptual disturbances, and any human-robot contact counts as failure. With the strongest policy, Xiaomi-Robotics-1, TRIP raises safety, the fraction of episodes without human contact among the scenes that can be solved without the person, from 61.3\% to 81.3\% relative to a turn-based harness using the same agent, action tools, and prompt, halving human contact. Task success is 38.7\% versus 31.0\% with the person and 73.2\% versus 73.8\% without, while agent invocations decrease from 30.3 to 19.4 per minute of robot time. TRIP is also safer than the turn-based harness with each of the other three policies. An ablation with Xiaomi-Robotics-1 shows that keeping the initial tripwire unchanged lowers safety to 61.3\%. On two real-robot tasks, hanging a mug and hanging a towel, TRIP succeeds in 25 of 40 trials across two disturbance conditions (blocking the arm and moving the target object), compared with 5 of 40 for the turn-based harness and 1 of 40 for the policy alone.

Our contributions are:
\begin{itemize}
\item \textbf{TRIP}, a training-free, policy-agnostic harness in which robot agents program their own interrupts: stateful tripwires that query a compact VLM during execution, compute on its answers, and are updated as the situation changes.
\item \textbf{RoboHD}, a human-disturbance benchmark of 168 scenes built on 42 RoboCasa365 tasks, where any human-robot contact counts as failure.
\item Experiments with three VLAs, one WAM, and a real robot: a matched comparison in which TRIP halves human contact with the strongest policy, an ablation on updating tripwires, and real-robot trials under disturbance.
\end{itemize}

%% file: sec/2_related.tex
\section{Related Work}
\label{sec:related}

\paragraph{Robot foundation models and agentic manipulation.}
Recent advances in robot foundation models have enabled increasingly general manipulation capabilities. Vision-language-action (VLA) models learn transferable policies from large-scale robot data~\cite{brohan2022rt,brohan2023rt,openx2024,kim2024openvla,octo2024,bousmalis2023robocat,black2024pi_0,intelligence2025pi_,bjorck2025gr00t,team2026xiaomi}, with diffusion-based architectures~\cite{chi2023diffusion,liu2024rdt,zhao2023learning} and multimodal conditioning~\cite{jiang2022vima} further expanding their capabilities. World action models extend this direction by jointly modeling actions and future observations~\cite{ye2026world,wang2026action}. Building on these pretrained models, language-model agents enable higher-level embodied reasoning~\cite{driess2023palme,ahn2022can}, program generation~\cite{liang2023code,singh2023progprompt,huang2023voxposer}, feedback-driven planning~\cite{huang2022inner,rana2023sayplan,zha2023droc,liu2023reflect}, and policy orchestration~\cite{zhang2026harnessvla,chen2026showharness,wang2026auriga,team2026systematically}. However, in turn-based agentic harnesses, the agent typically regains control only after a tool call returns, limiting its ability to respond to changes during execution. TRIP addresses this limitation by allowing the agent to interrupt ongoing policy execution without modifying the underlying policy.

\begin{figure}[t]
\centering
\includegraphics[width=\linewidth]{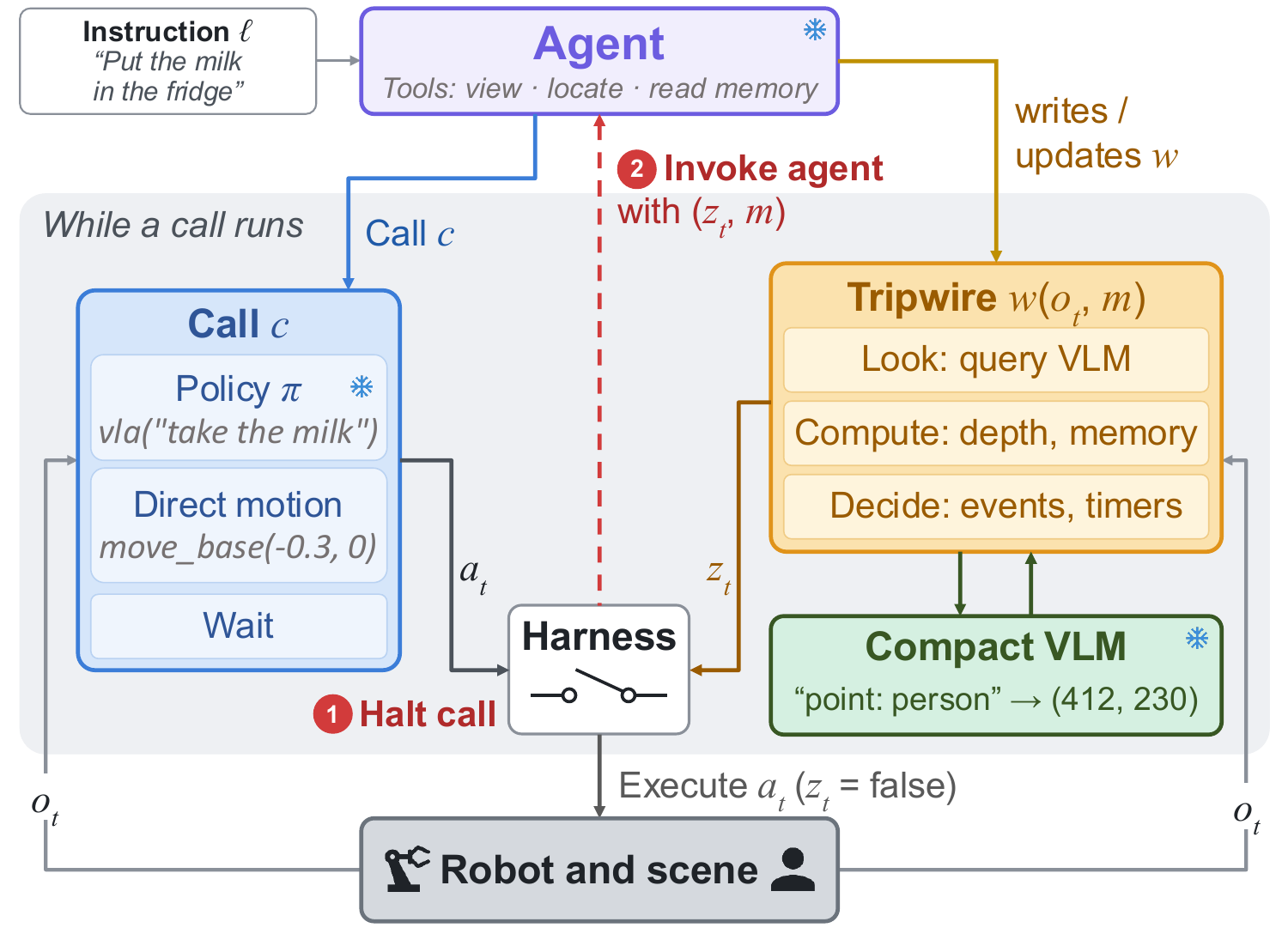}
\caption[TRIP]{\textbf{Overview of TRIP.} Given an instruction $\ell$, the agent selects a call $c$ and a tripwire $w$. During execution, the tripwire reads observations $o_t$, queries a compact VLM, and combines the results with depth and memory $m$ to produce $z_t$. A false output allows execution to continue; a firing stops the call and returns control to the agent with the reason and memory. The agent then reassesses the scene and updates the call and tripwire. Snowflakes denote frozen models; calls and VLM queries are illustrative.}
\label{fig:method}
\end{figure}

\paragraph{Runtime monitoring and event-driven control.}
Execution monitoring has been studied through language constraints, visual feedback, and generated programs~\cite{guo2024doremi,zhou2025code,xu2025can}. DoReMi~\cite{guo2024doremi} detects plan-execution misalignment through visual checks, while Code-as-Monitor~\cite{zhou2025code} evaluates generated spatiotemporal constraints. VoLo~\cite{chen2026volo} periodically monitors running policies with a VLM, and The Harness~\cite{xu2026harness} uses agent-written reactive guards. Event-triggered control similarly updates decisions when specified conditions hold~\cite{tabuada2007event,heemels2012introduction}. TRIP enables agents to write and revise stateful interruption programs that combine on-demand VLM perception with depth, geometry, and temporal memory, determining when to return control to the agent.

\paragraph{Human-robot interaction benchmarks.}
Manipulation benchmarks evaluate generalization across tasks, objects, and environments~\cite{james2020rlbench,mees2022calvin,gu2023maniskill2,mandlekar2021robomimic,pumacay2024colosseum,fei2026libero,zhou2025libero}, while recent work studies disturbances during execution~\cite{zhang2026libero,fang2026towards}. Embodied household environments~\cite{shridhar2020alfred,li2023behavior,padmakumar2022teach,puig2018virtualhome,li2022igibson} and human-aware embodied environments and benchmarks~\cite{puig2024habitat,chang2025partnr,zhou2026harp} investigate interactive tasks and human-centered settings. RoboHD extends RoboCasa365~\cite{nasiriany2026robocasa365} with robot-state-triggered human disturbances in mobile manipulation, evaluating both task completion and strict human-robot no-contact safety.

%% file: sec/3_method.tex
\section{Method}
\label{sec:method}

TRIP is a training-free harness for robot manipulation under human disturbance.
Its vision-language agent chooses between policy execution and direct robot control, and attaches to each call a \emph{tripwire}, a stateful Python program that monitors the running call and determines when to return control to the agent.
We describe the execution loop, tripwires, agent tools, and RoboHD in \cref{sec:overview,sec:tripwire,sec:agent,sec:dataset}.

\begin{algorithm}[t]
\caption{Execution Loop of TRIP}
\label{alg:loop}
\begin{algorithmic}[1]
\Require Policy $\pi$, instruction $\ell$, agent $\mathcal{A}$
\State $(c, w) \gets \mathcal{A}(\ell)$; \quad $m \gets \varnothing$
\While{$c \neq \texttt{finish}$}
  \State $o_t \gets \Call{Observe}{}$
  \State $(m, z_t) \gets w(o_t, m)$
    \label{alg:trip} \Comment{Evaluate tripwire}
  \If{$z_t \neq \texttt{false}$}
    \State \Call{StopAndHold}{$c$}
    \State $(c, w) \gets \mathcal{A}(\ell, z_t, m)$; \quad
      $m \gets \varnothing$
      \label{alg:agent}
  \Else
    \If{$c \text{ is a policy call}$}
      \State $a_t \gets \pi(o_t, c)$
    \Else
      \State $a_t \gets \kappa_c(o_t)$
        \Comment{Direct motion or hold}
    \EndIf
    \State \Call{Execute}{$a_t$}
  \EndIf
\EndWhile
\end{algorithmic}
\end{algorithm}

\subsection{Event-driven Control}
\label{sec:overview}

Given a task instruction $\ell$, the agent $\mathcal{A}$ selects an execution call $c$ and a tripwire $w$ (\cref{fig:method,alg:loop}).
The call runs either a pretrained manipulation policy, $a_t = \pi(o_t,c)$, or a direct command, $a_t = \kappa_c(o_t)$, which moves the base, arm, or gripper, or holds the robot still; the policy's parameters and architecture stay unchanged.
During execution, the tripwire checks observations and updates its memory $m$.
If it returns \texttt{false}, execution continues.
Otherwise, it returns a firing reason $z_t$, and the harness stops the call and holds the robot still before waking the agent, so the stop does not wait for agent inference.
When a call ends on its own, such as a completed base motion, the robot holds still until the tripwire, which sees that the call is over, fires.

The agent receives the reason, the tripwire's memory, and current camera images, and issues the next call; we refer to this decision as an \emph{agent turn}.
With the call, it can retain or update the tripwire to match its chosen response.
For example, during manipulation the tripwire may watch for a person approaching; after the agent decides to wait, it may instead watch for the person leaving (\cref{fig:teaser}).
The agent thereby determines what change should prompt its next turn.

\subsection{Programmable Tripwires}
\label{sec:tripwire}

A tripwire is a Python function
\begin{equation}
    (m_{t+1}, z_t) = w(o_t, m_t),
\end{equation}
where $m_t$ stores information across checks and $z_t$ is either \texttt{false} or a textual firing reason, such as \emph{``person 0.6\,m away, closing at 0.3\,m/s''}, which gives the agent the evidence behind the interruption.
The observation $o_t$ includes RGB-D images, robot masks, gripper and base poses, and execution metadata: the active call, elapsed times, and the time and reason of the previous firing.

\paragraph{Visual perception and computation.}
The tripwire accesses a compact, frozen VLM through \texttt{vlm(question, images)}, choosing the camera views, questions, and query frequency.
It combines the VLM's short responses with geometric measurements and temporal information: a predicted image location and depth give a person's position relative to the robot, and successive estimates show whether the person is approaching.
With timers, step counts, and logical conditions, a single tripwire can monitor task progress, unexpected arrivals, or displacement of a target object.

\paragraph{Temporal memory and re-arming.}
The tripwire's bounded memory holds recent distances, observations, or counters.
When the tripwire fires, this memory is passed to the agent and then cleared for the next call (line~\ref{alg:agent} of \cref{alg:loop}).
To carry information into the next call, the agent can embed it in the tripwire code as constants; together with the previous firing's time and reason, this lets the next tripwire avoid firing repeatedly while the same condition holds, for instance by firing only if a nearby person comes closer.

\paragraph{Bounded execution.}
The harness runs each tripwire in a sandbox with restricted imports, no direct file or network access, and limits on code size, memory, VLM queries, and execution time (\cref{sec:env}).
If a check raises a runtime error, exceeds its time limit, or returns a malformed output, the harness stops the active call and wakes the agent with the error so that it can revise the program.

\subsection{Agent Decisions and Tools}
\label{sec:agent}

The agent is a vision-language model with tool-calling support.
At each turn, it receives the task instruction and the episode history: previous observations, reasoning, tool calls and results, and tripwire programs and firings (omitted from the arguments of $\mathcal{A}$ in \cref{alg:loop}).
It has five tool groups:
\begin{itemize}
\item \textbf{Observation:} inspect camera images and recover 3D locations from selected pixels and depth.
\item \textbf{Policy execution:} run $\pi$ with a language instruction, using a tripwire to determine when to interrupt rather than choosing a fixed duration for each call.
\item \textbf{Direct action:} move the base, arm, or gripper, grasp at a point, or hold still.
\item \textbf{Tripwire programming:} write a Python monitor that replaces the active tripwire.
\item \textbf{Memory inspection:} inspect the memory returned with a firing, such as the sequence of distances preceding it.
\end{itemize}
Within a turn, the agent can inspect observations and tripwire memory while the robot remains stationary, update the tripwire, and then issue a policy or direct-action call, such as resuming manipulation, waiting for the workspace to clear, repositioning the robot, or recovering a displaced object.
It terminates the episode with \texttt{finish}.

%% file: sec/3_dataset.tex
\subsection{RoboHD: Human Disturbance Benchmark}
\label{sec:dataset}

We build \textbf{RoboHD}, a benchmark for robot manipulation under human disturbance.
It extends RoboCasa365~\cite{nasiriany2026robocasa365} with a reactive human avatar that shares the kitchen with the robot and can interfere with its task; an episode succeeds only if the task is completed without human--robot contact.

\begin{figure*}[t]
\centering
\includegraphics[width=\textwidth]{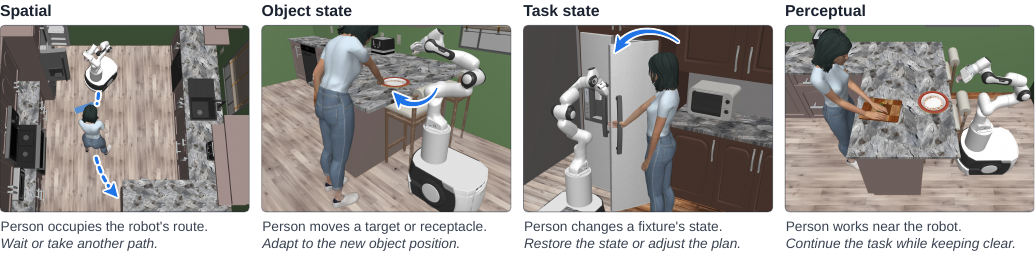}
\caption{\textbf{Disturbance types in RoboHD.} One scene of each type; each type has 42 scenes. Below each picture, the upper line describes the human disturbance and the lower line describes the robot's required response.}
\label{fig:robohd}
\end{figure*}

\paragraph{Scene construction.}
We select 42 composite tasks from RoboCasa365, each requiring the
PandaOmron mobile manipulator to navigate between fixtures, allowing
human activity to interfere with both navigation and manipulation.
For each task, we construct four RoboHD scenes, one per disturbance
type, yielding 168 scenes in total.
Each scene preserves the kitchen, objects, and instruction of its
corresponding RoboCasa365 task and adds a human avatar at an assigned
activity location.
Of the 168 scenes, 36 are derived from 9 tasks in the target split
and 132 from 33 tasks in the pretrain split.
For each scene, we use the first initialization seed that accommodates
the avatar in the kitchen.
In 113 scenes, the avatar's activity location lies within 0.9\,m of
a base position visited during task execution without the person,
with a median distance of 0.61\,m.
All methods use the same initial state for each scene, and we also
evaluate each scene without the person as a reference.

\paragraph{Disturbance types.}
The four disturbance types are illustrated and summarized in \cref{fig:robohd}.
\emph{Spatial} disturbances obstruct the robot's navigation path: a person stands or performs an activity such as mopping along the route, requiring the robot to wait or take another path.
\emph{Object-state} disturbances change the position of an object needed for the task, such as the next grasp target or a placement receptacle, and the robot must adapt to the new position.
\emph{Task-state} disturbances change the state of a fixture, such as closing a refrigerator, opening a cabinet, or turning on a faucet or burner; the robot must restore the required state or adjust its plan.
\emph{Perceptual} disturbances leave task-relevant objects unchanged: the person performs an unrelated activity, such as chopping, grating, or peeling, within 1\,m of the robot and in its view, and the robot must continue the task while keeping clear of the person.

\paragraph{Reactive human simulation.}
Each scene specifies an avatar activity and a location near the corresponding object, fixture, or navigation path.
A scripted controller executes the activity using motion-capture clips from BONES-123~\cite{BONES} and SEED~\cite{SEED}, chaining walking, turning, and idle clips to move the avatar through the kitchen and blending action sequences for compound motions such as walking while sweeping.

The controller follows five stages, \emph{idle, approach, wait, act,} and \emph{leave}, and advances to the next when the robot reaches a triggering state: grasping or releasing an object, the gripper or base approaching a target, or a door reaching a specified angle.
For route-blocking scenes, the avatar steps away once the robot comes within 0.5\,m.
The avatar therefore reacts to task progress rather than following a fixed time schedule, and can disturb the robot while a policy call is still running.

Because the avatar has no collision body in the simulator, the contact check is geometric: we fit capsules to the avatar's current skeleton, represent the robot using bounding boxes of its collision geometries, and count a minimum capsule--box distance of zero or less as contact.
Any contact fails the episode, regardless of whether the task could otherwise be completed.

%% file: sec/4_experiments.tex
\section{Experiments}
\label{sec:experiments}

We evaluate TRIP, wrapped around several pretrained policies used as is, on RoboHD (\cref{sec:dataset}), in which touching the person fails the episode (\cref{sec:baselines,sec:eval}).
The policy alone shows what TRIP adds; an agentic harness baseline with TRIP's agent, tools and prompt isolates what TRIP changes: its agent chooses how long each policy call runs and is invoked again only when the call ends, while TRIP's agent is woken when a tripwire fires.
We also run TRIP on a real robot (\cref{sec:real}).

\input{tables/main}

\subsection{Setup}
\label{sec:env}

The robot (\cref{sec:dataset}) is controlled at 20\,Hz.
The agent sees through five RGB-D cameras: the robot's two front cameras and wrist camera, which the policies also use, and two wide-angle cameras we mount on its back, facing rear-left and rear-right, so that it can look in every direction by calling for any camera's picture.

An episode of an agentic method ends when the agent calls \texttt{finish}, at the first contact with the person, or once it reaches its spend cap or 100 tool calls. Unless otherwise stated, the agent is GPT-6.1-sol, and the physics pauses while it or the VLM computes, which is common practice for agentic manipulation in simulation, where a frontier agent takes seconds per decision~\cite{tsui2026faea, fu2026cap, zhang2026harnessvla, team2026systematically}.
Nothing pauses in the real-robot experiments (\cref{sec:real}).

In TRIP, the tripwires query Qwen3.5-9B~\cite{qwen2026qwen35} with its thinking turned off.
A tripwire is stepped at every control step and queries the VLM about twice per second of robot time (\cref{sec:ablation}).
A firing tripwire halts the policy at that step, even in the middle of an action chunk.
Until the agent writes its first tripwire, a default one fires when the call ends or after 8\,s.
A tripwire works within the following limits:
\begin{itemize}
\item \textbf{Code.} It runs in a sandbox that allows only the numpy, math, json, re and statistics modules, with no file access, and its source has at most 6000 characters; a source that breaks these rules is refused when the agent writes it.
\item \textbf{VLM.} Each check may query the VLM at most twice, for answers of at most 64 tokens.
\item \textbf{Time.} Each check must return within 4\,s. If the tripwire stays silent, a policy call stops after 60\,s, and the agent is woken once the robot has held still for 10\,s.
\item \textbf{Memory.} Its memory holds at most as much as two frames of every camera's image and depth, and starts empty at each agent turn.
\end{itemize}

\subsection{Baselines}
\label{sec:baselines}

We compare TRIP with two kinds of baselines: pretrained policies run alone, and a turn-based agentic harness baseline that wraps the same policies as TRIP.

\paragraph{Policies alone.}
We run four open-source policies on their own, three VLAs and a WAM, each trained on RoboCasa365 data.
The VLA GR00T N1.5~\cite{bjorck2025gr00t} is a baseline released with RoboCasa365~\cite{nasiriany2026robocasa365}; we use its multitask checkpoint on all scenes.
The VLAs REKI-2B~\cite{marrie2026reki} and Xiaomi-Robotics-1 (XR)~\cite{team2026xiaomi} and the WAM PatchWAM~\cite{wang2026action} were among the strongest entries on its leaderboard when they were released.

\paragraph{Agentic harness baseline.}
The agentic harness baseline (Agentic in \cref{tab:calls,tab:real}) follows the setup of RPent~\cite{zhang2026harnessvla}: its agent acts only through tool calls, is invoked again after each one, and ends the episode with \texttt{finish}.
Its tools are TRIP's observation and direct-action tools (\cref{sec:agent}) and a policy call that runs for a duration the agent chooses, from 1 to 16\,s, and stops early only on success.
TRIP uses the same cameras and tools, with a tripwire in place of the chosen duration, and the same prompt except for the part on tripwires, so the two methods differ only in when the agent is invoked.
Other harnesses do not consider a person who interrupts the robot, and fixed-interval checks~\cite{chen2026volo} are covered by this baseline, whose agent looks at the scene about every 2\,s (\cref{tab:calls}).

\subsection{Evaluation}
\label{sec:eval}

\paragraph{Metrics.}
We report three numbers per method:
\begin{itemize}
\item \textbf{Success with the person.} The task's own check passes and the robot never touches the person.
\item \textbf{Success without the person.} The task's own check passes.
\item \textbf{Safety.} The share of episodes with the person without contact, counted for each policy over the scenes that at least one of the three methods (the policy alone, the agentic harness baseline or TRIP) solves without the person, so that a policy is not counted as safe in scenes where it never gets far into the task.
\end{itemize}

\begin{figure}[t]
\centering
\includegraphics[width=\linewidth]{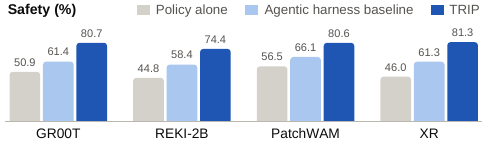}
\caption[Safety]{Safety (\%) with the person, for each policy alone, with the agentic harness baseline and with TRIP, over the scenes that at least one of the three methods solves without the person (57 scenes for GR00T, 125 for REKI-2B, 124 for PatchWAM and 150 for XR).}
\label{fig:safety}
\end{figure}

\paragraph{Overall performance.}
Results are shown in \cref{tab:main,fig:safety}.
With the person, TRIP raises success over the policy alone with GR00T, REKI-2B, PatchWAM and XR.
It also succeeds more often than the agentic harness baseline with REKI-2B, PatchWAM and XR; with XR it is the best method overall, raising the success of the policy alone from 20.8\% to 38.7\%.
Since the two agentic methods differ only in when the agent is invoked, we attribute the gain over the baseline to the tripwires, which watch the person throughout each call rather than only between tool calls.

\paragraph{With and without the person.}

The performance of both the agentic harness baseline and TRIP rests on the policy each wraps. Without disturbance, the two agentic methods lift each policy by similar margins, which depend only on the policy itself.
For TRIP, the share of its success that survives the person grows with the strength of the policy, from 39\% with GR00T to 53\% with XR.

\paragraph{Safety.}
\cref{fig:safety} shows the safety of every method, grouped by policy. We see safety is higher with TRIP regardless of the low-level policy, which we attribute to TRIP actively avoiding contacts through direct robot control. Among all methods, TRIP with XR achieves the highest safety performance.

\subsection{Ablation}
\label{sec:ablation}

All numbers in this subsection are for TRIP and the agentic harness baseline with XR on the 168 RoboHD scenes with the person.

\input{tables/ablation}

\paragraph{Fixed tripwire.}
We run TRIP with a tripwire that stays fixed for the whole episode (\cref{tab:ablation}): either the one the agent writes on its first turn, or a pre-defined one that holds the robot still while the person is within 0.6\,m and wakes the agent once the person has stayed beyond 0.7\,m for 1\,s.
The agent-designed tripwire lowers safety from 81.3\% to 61.3\% and success from 38.7\% to 33.3\%; the pre-defined one reaches the highest safety (86.0\%), mostly by holding the robot still, but its success drops to 16.1\%.
TRIP therefore needs to rewrite its tripwire.

\paragraph{Agentic model.}
We test GPT-6 Astra~\cite{openai2026astra} as TRIP's agent for XR, with results shown in \cref{tab:ablation}. We do not test all methods with Astra due to high API costs. Across models, success is unchanged (38.1\% vs.\ 38.7\%); safety is lower (73.8\% vs.\ 81.3\%) but stays above that of the agentic harness baseline with GPT-6.1-sol (61.3\%), showing no significant improvement from changing the agentic model.

\input{tables/calls}

\begin{figure*}[t]
\centering
\includegraphics[width=\textwidth]{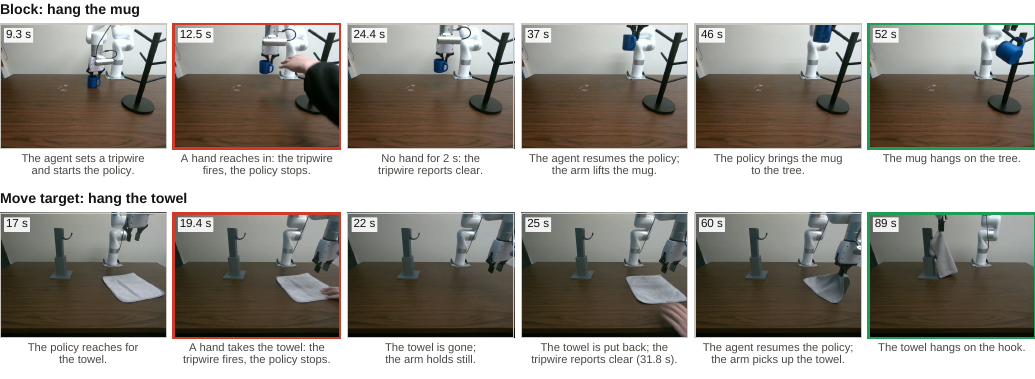}
\caption[TRIP on the real robot]{TRIP in two real-robot episodes (front camera). Top, block: a hand reaches toward the arm while it hangs the mug; the tripwire stops the policy, and once no hand has been seen for 2\,s the agent resumes it and the mug is hung. Bottom, move target: a hand takes the towel; the tripwire stops the policy and the arm holds still until the towel is back; then the agent resumes the policy and the towel is hung.}
\label{fig:real_episode}
\end{figure*}

\paragraph{Agent call analysis.}
TRIP's agent makes 19.4 tool calls per minute of robot time, against 30.3 for the agentic harness baseline (\cref{tab:calls}).
The baseline spends most of its tool calls looking: 28.7 view or locate calls per episode, to check on the person between policy calls, against 10.1 for TRIP, whose tripwires do the watching; they fire 10.3 times per minute and ask the VLM 2.2 times per second of robot time.
The agent writes 4.2 tripwires per episode, the most in object-state scenes (4.8), where the person moves what the robot handles.

\subsection{Real-world Experiments}
\label{sec:real}

\paragraph{Hardware.}
We use a UFACTORY xArm 7, a 7-DoF arm with a parallel gripper (\cref{fig:real_episode}).
Two Intel RealSense D455 cameras observe the workspace: a fixed third-person camera and a wrist camera.

\paragraph{Policies.}
The base policies are two single-task $\pi_{0.5}$~\cite{intelligence2025pi_} policies, one that hangs a mug on a mug tree and one that hangs a towel. These policies are trained with behavior cloning, each using 100 demonstrations collected with teleoperation.
Each takes the RGB images of both cameras and the proprioceptive state, and predicts target end-effector poses and gripper commands, which are executed at 30\,Hz.
The policy runs on one RTX 4090; under TRIP, a second RTX 4090 runs Qwen3.5-9B, the VLM the tripwires query.

\paragraph{Protocol.}
We test both tasks, each in 10 different table setups.
In each setup we run the policy alone, the agentic harness baseline and TRIP under three conditions:
\begin{itemize}
\item \textbf{Original.} The task as it is: nobody enters the workspace.
\item \textbf{Block.} A person blocks the arm's path; an episode in which the robot touches the person fails.
\item \textbf{Move target.} A person moves the mug or the towel away while the robot works on it and puts it back after a while.
\end{itemize}
An episode succeeds when the mug hangs on the mug tree or the towel is hung up; with the person, only if the robot never touches the person.
In every trial the person is one of the authors, who consented to take part, so we sought no IRB review.
\Cref{tab:real} gives the successes over the 10 setups of each task, and \cref{fig:real_episode} shows one TRIP episode of each disturbance.

\input{tables/real}

\paragraph{Agents.}
The agentic harness baseline and TRIP wrap the same $\pi_{0.5}$ policy, with GPT-6.1-sol as the agent.

\paragraph{Results.}
Without disturbance the three methods are close (5--7 of the 10 setups), but when a person blocks the arm or moves the target object, the policy alone and the agentic harness baseline succeed in at most 3 setups, while TRIP succeeds in 5--7, about as often as without disturbance (\cref{tab:real}).

\paragraph{Latency.}
Nothing pauses on the real robot: a tripwire check takes 0.24\,s on average (median 0.22\,s), including 1.4 VLM queries per check of 0.17\,s each.
The checks run alongside the policy on each new picture, so the tripwire reads about four pictures per second.
This rate is below the policy's 30\,Hz control rate but matches the frequency of its action chunks, and is fast enough to detect an anomaly on the real robot. An agent turn takes 4.3\,s on average (median 3.2\,s), during which the arm stands still.

%% file: tables/main.tex
\begin{table*}[t]
\centering
\footnotesize
\setlength{\tabcolsep}{6pt}
\begin{tabular}{@{}llcccccc@{}}
\toprule
& & \multicolumn{5}{c}{With the person} & Without \\
\cmidrule(lr){3-7}\cmidrule(l){8-8}
Policy & Method & Spatial & Object state & Task state & Perceptual & All & All \\
\midrule
\multirow{3}{*}{GR00T} & Policy alone & 4.8 & 2.4 & 0.0 & 4.8 & 3.0 & 10.1 \\
 & Agentic harness baseline & \textbf{7.1} & \textbf{7.1} & 2.4 & \textbf{7.1} & 6.0 & \textbf{18.5} \\
 & TRIP (ours) & \textbf{7.1} & 4.8 & \textbf{7.1} & \textbf{7.1} & \textbf{6.5} & 16.7 \\
\midrule
\multirow{3}{*}{REKI-2B} & Policy alone & 9.5 & 11.9 & \textbf{14.3} & 19.0 & 13.7 & 37.5 \\
 & Agentic harness baseline & 21.4 & 9.5 & 11.9 & 19.0 & 15.5 & 51.2 \\
 & TRIP (ours) & \textbf{31.0} & \textbf{23.8} & \textbf{14.3} & \textbf{21.4} & \textbf{22.6} & \textbf{53.6} \\
\midrule
\multirow{3}{*}{PatchWAM} & Policy alone & 16.7 & 11.9 & 21.4 & \textbf{23.8} & 18.5 & 39.9 \\
 & Agentic harness baseline & 21.4 & \textbf{16.7} & 16.7 & 19.0 & 18.5 & 53.0 \\
 & TRIP (ours) & \textbf{31.0} & \textbf{16.7} & \textbf{26.2} & \textbf{23.8} & \textbf{24.4} & \textbf{56.5} \\
\midrule
\multirow{3}{*}{XR} & Policy alone & 33.3 & 7.1 & 19.0 & 23.8 & 20.8 & 61.9 \\
 & Agentic harness baseline & 33.3 & 23.8 & 28.6 & \textbf{38.1} & 31.0 & \textbf{73.8} \\
 & TRIP (ours) & \textbf{50.0} & \textbf{26.2} & \textbf{40.5} & \textbf{38.1} & \textbf{38.7} & 73.2 \\
\bottomrule
\end{tabular}
\caption[Success on the 168 scenes]{Success rate (\%) on the 168 RoboHD scenes (42 per disturbance category), for four policies run alone, wrapped by the agentic harness baseline, and wrapped by TRIP.
With the person, an episode succeeds only if the task succeeds and the robot never touches the person.
The agent is GPT-6.1-sol. Bold: the best of the three methods for each policy.}
\label{tab:main}
\end{table*}

%% file: tables/ablation.tex
\begin{table}[t]
\centering
\footnotesize
\setlength{\tabcolsep}{4pt}
\begin{tabular}{@{}llcc@{}}
\toprule
Tripwire & Variant & Success & Safety \\
\midrule
\multirow{2}{*}{Dynamic} & Original TRIP & 38.7 & 81.3 \\
 & GPT-6 Astra as the agent & 38.1 & 73.8 \\
\midrule
\multirow{2}{*}{Fixed} & Agent-designed & 33.3 & 61.3 \\
 & Pre-defined & 16.1 & 86.0 \\
\bottomrule
\end{tabular}
\caption[Ablations of TRIP with XR]{TRIP with XR on the 168 scenes with the person, against three variants. Dynamic: the agent can rewrite its tripwire at any turn, with GPT-6.1-sol (TRIP) or GPT-6 Astra as the agent. Fixed: one tripwire for the whole episode, either designed by the agent on its first turn or pre-defined, holding the robot still while the person is within 0.6\,m. Success (\%) is over the 168 scenes; safety (\%) is over the 150 scenes of XR, as in \cref{fig:safety}.}
\label{tab:ablation}
\end{table}

%% file: tables/calls.tex
\begin{table}[t]
\centering
\footnotesize
\setlength{\tabcolsep}{5pt}
\begin{tabular}{@{}lcccc@{}}
\toprule
& \multicolumn{2}{c}{Tool calls per minute} & \multicolumn{2}{c}{Looks per episode} \\
\cmidrule(lr){2-3}\cmidrule(l){4-5}
Disturbance & TRIP & Agentic & TRIP & Agentic \\
\midrule
Spatial & 19.4 & 39.8 & 10.7 & 31.4 \\
Object state & 20.6 & 25.2 & 10.9 & 26.1 \\
Task state & 18.7 & 25.8 & 9.3 & 23.8 \\
Perceptual & 18.9 & 32.7 & 9.5 & 33.6 \\
\midrule
All & 19.4 & 30.3 & 10.1 & 28.7 \\
\bottomrule
\end{tabular}
\caption[Agent calls]{Agent tool calls per minute and look calls (view, locate) per episode, with XR.}
\label{tab:calls}
\end{table}

%% file: tables/real.tex
\begin{table}[t]
\centering
\footnotesize
\setlength{\tabcolsep}{3.5pt}
\begin{tabular}{@{}llccc@{}}
\toprule
Task & Method & Original & Block & Move target \\
\midrule
\multirow{3}{*}{Mug} & $\pi_{0.5}$ alone & 5 & 0 & 0 \\
 & Agentic -- $\pi_{0.5}$ & 6 & 0 & 2 \\
 & TRIP (ours) -- $\pi_{0.5}$ & 6 & \textbf{6} & \textbf{5} \\
\midrule
\multirow{3}{*}{Towel} & $\pi_{0.5}$ alone & 7 & 0 & 1 \\
 & Agentic -- $\pi_{0.5}$ & 6 & 0 & 3 \\
 & TRIP (ours) -- $\pi_{0.5}$ & 7 & \textbf{7} & \textbf{7} \\
\bottomrule
\end{tabular}
\caption[Real-robot successes]{Successes on the real robot over the 10 table setups of each task. Touching the person fails the episode.}
\label{tab:real}
\end{table}

%% file: sec/7_conclusion.tex
\section{Limitations}
\label{sec:limitations}

\begin{itemize}
\item The person is simulated: an avatar that plays motion-capture clips under a scripted controller, with no collision body. Large-scale real-world tests with human subjects are infeasible because unexpected robot behavior is potentially harmful to the subjects.
\item The physics pauses while the agent and the tripwire's VLM compute, so the robot's time excludes their latency. A TRIP episode with GPT-6.1-sol costs \$0.2--0.3 in API calls and takes about 12\,min of wall-clock time for 1.5\,min of robot time.
\item Due to high (10$\times$) API costs, only Xiaomi-Robotics is tested with the current state-of-the-art agentic model, GPT-6 Astra. We expect future, more powerful models to show stronger manipulation abilities on RoboHD.
\end{itemize}

\section{Conclusion}
\label{sec:conclusion}

TRIP lets the agent program its own interrupts: it writes tripwires that watch the scene while the policy acts, so the agent is no longer blind until a tool call returns, and the policy is used as is, without training.
On RoboHD, our benchmark of RoboCasa365 scenes with a reactive human avatar, TRIP halves human contact with the strongest policy relative to a turn-based harness with the same agent, tools and prompt, raising safety from 61.3\% to 81.3\% without lowering task success.  On a real robot, TRIP succeeds under human disturbance about as often as without it, where the turn-based harness and the policy alone largely fail. Our experiments show that TRIP is an effective agentic manipulation framework in human-disturbed scenes, which may reduce the need for data from such scenes in training the low-level policies.

%% file: sec/X_suppl.tex
\clearpage
\maketitlesupplementary

This supplement gives the implementation of TRIP and of the baselines with two tripwires written by the agent (\cref{sec:impl}), further results, including a run with Claude Opus 5.5 as the agent (\cref{sec:more}), and details of RoboHD (\cref{sec:robohd_detail}).

\section{Implementation details}
\label{sec:impl}

\subsection{Agent and tools}
\label{sec:tools}

The agent, GPT-6.1-sol, runs with reasoning effort xhigh in simulation and medium on the real robot.
An episode is one conversation.
At each turn the agent receives the outputs of its last tool calls, the robot time and wall-clock time used, the money spent, the gripper's measured position and the base's odometry.
It may issue several tool calls in one reply; they run in order, and a firing tripwire cancels the calls still to run.
Camera pictures are $256\times256$ pixels, and none arrives unasked: the agent sees the scene through \texttt{view}, and, in TRIP, through the pictures of the tripwire's cameras at the step it fired.
The spend cap is \$0.40 per episode.

\input{tables/supp_tools}

\Cref{tab:supp_tools} lists the tool calls.
Memory inspection needs no call of its own: when a tripwire fires, the output of the interrupted call gives the reason, the tripwire's memory at that step (numbers and strings in full, arrays by their shape) and the pictures of the tripwire's cameras.
Every output of \texttt{locate} also gives each point in the kitchen's fixed frame, so that a tripwire can keep the distance to a located person or object while the robot moves.

\subsection{Prompts}
\label{sec:prompt}

Both agentic methods use one system prompt for all 42 tasks and all four disturbance types; the task instruction is the first user message.
The prompt has the following parts:
\begin{itemize}
\item the task, the tool calls and the cameras;
\item the person: what they may be doing and when they are in the robot's way;
\item the coordinate frames;
\item rules for acting;
\item strategy priors;
\item in TRIP only: the tripwire interface and its limits (\cref{sec:interface}), a template tripwire (\cref{sec:bench}), and when a tripwire should fire.
\end{itemize}
The parts on the task, the cameras, the person and the strategy priors are the same in both methods; the baseline's rules say for a turn-based agent what TRIP's rules say for tripwires.
The prompt does not say which disturbance type an episode has or when the disturbance happens.

\subsection{Tripwire interface}
\label{sec:interface}

A tripwire is a Python function \texttt{monitor(state, memory)} that returns \texttt{(memory, trigger)}.
The harness calls it at every control step (0.05\,s of robot time) while a tool call runs and while the robot holds still after one.
A \texttt{trigger} that is false lets the robot go on; a string fires the tripwire and is the reason the agent is given.
\texttt{state} holds:
\begin{itemize}
\item \texttt{t}, \texttt{t\_since\_turn}: the robot time and the time since the agent's last turn;
\item \texttt{images}, \texttt{depth}: the picture and the depth image of each camera the agent named for this tripwire;
\item \texttt{robot}: for each camera, a mask of the pixels that show the robot's own arm, gripper and base;
\item \texttt{gripper}, \texttt{fingers\_m}, \texttt{base}: the gripper's position in the robot's frame, the opening of its fingers, and the base's odometry in the kitchen's fixed frame;
\item \texttt{tool}, \texttt{tool\_t}, \texttt{tool\_done}: the running call, how long it has run, and whether it is over;
\item \texttt{last\_trigger}: how long ago the agent's tripwire last fired, and its reason.
\end{itemize}
The function may call \texttt{vlm(question, images, max\_tokens)} and \texttt{to\_robot(point, base)}, which turns a point of the kitchen's fixed frame into the robot's frame.
\texttt{memory} is a tuple of arrays, strings and numbers; it is \texttt{None} at the first step after each agent turn.
The limits are those of \cref{sec:env}; the memory limit, two frames of every camera's picture and depth, is 4.6\,MB.

\subsection{The template tripwire}
\label{sec:bench}

TRIP's prompt contains a template tripwire, which the agent adapts at each turn.
Its helper \texttt{person\_depth} asks the VLM to point at the person in the two front views and in the two rear views, each pair stitched side by side into one $1024\times512$ picture, so that the two VLM queries allowed per check cover four cameras; it reads the depth at the point the VLM gives.
\texttt{monitor} samples it every 0.8\,s and fires when the person is within a margin, when they close at more than 0.3\,m/s within 1.5\,m, or after 8\,s for a progress check.
The margin is 0.6\,m while the person moves and 0.3\,m once they have stood still for 5\,s, so that the robot can work beside a person who stays at their own work.
After a firing on the person the template re-arms from that firing and fires again only when the person is 0.15\,m nearer, since a condition that already holds when the memory restarts would otherwise fire at every sample.

\subsection{Policies and episodes}
\label{sec:policies}

Each policy is served as in its own RoboCasa365 evaluation and sees the robot's two front cameras and wrist camera, never the rear cameras.
\begin{itemize}
\item \textbf{GR00T N1.5}: the multitask checkpoint, with 4 denoising steps; each 16-step action chunk is executed in full.
\item \textbf{REKI-2B}: its RoboCasa365 checkpoint; one prediction of 20 actions per second of robot time, executed in full.
\item \textbf{XR}: its RoboCasa365 checkpoint; the input is a window of 4 frames per camera, 2 steps apart, and each 16-step chunk is executed in full.
\item \textbf{PatchWAM}: its RoboCasa365 checkpoint; each 16-step chunk is executed in full.
\end{itemize}
Under both agentic methods a policy call starts with a new chunk and executes chunks in the same way; a firing tripwire halts the policy at that step and the rest of the chunk is dropped.

A policy alone runs for the task's RoboCasa365 horizon.
For every method the task's own check is evaluated at every step, and the episode ends when it passes.

\subsection{Tripwires written by the agent}
\label{sec:examples}

\begin{figure}[t]
\fontsize{5.6}{6.5}\selectfont
\begin{verbatim}
import math
import re
import numpy as np
BODY=((3.96,-5.045,1.35),(3.883,-5.078,1.018),(3.867,-5.049,.935))
LEGS=((3.965,-5.006),(3.886,-4.946),(3.98,-4.915))
def monitor(state,memory):
    bx,by,yaw=state['base']
    co,si=math.cos(yaw),math.sin(yaw)
    def xy(x,y):
        dx,dy=x-bx,y-by
        return co*dx+si*dy,-si*dx+co*dy
    gx,gy,gz=state['gripper']
    arm=min(math.sqrt((gx-xy(x,y)[0])**2+(gy-xy(x,y)[1])**2
                      +(gz-z)**2) for x,y,z in BODY)
    edge=min(math.hypot(max(-.55-xy(x,y)[0],0,xy(x,y)[0]-.15),
                        max(abs(xy(x,y)[1])-.25,0)) for x,y in LEGS)
    if memory is None: memory=(state['t']-1.,edge,None)
    last,initial,prior=memory
    recovery=state['tool']=='move_base' and edge>=initial-.01
    if arm<.31 or (edge<.30 and not recovery):
        return memory,(f'person clearance: gripper {arm:.2f} m, '
                       f'base edge {edge:.2f} m')
    if state['t']-last>=1.:
        img=np.concatenate((state['images']['left'],
                            state['images']['right']),axis=1)
        said=vlm('Is the woman leaning or reaching her hand toward '
                 'the white robot gripper, rather than standing '
                 'with hands down? Reply only ANSWER: yes or '
                 'ANSWER: no.',[img],32)
        if 'ANSWER: yes' in said:
            return (state['t'],initial,prior),(
                f'woman reaching toward arm; gripper clearance '
                f'{arm:.2f} m, base {edge:.2f} m')
        last=state['t']
    return (last,initial,prior),(
        f'progress: gripper {arm:.2f} m, base edge {edge:.2f} m'
        if state['t_since_turn']>=8 or state['tool_done'] else False)
\end{verbatim}
\caption{A tripwire written by the agent, with line breaks added (AddLemonToFish, object state; the episode succeeds). \texttt{BODY} and \texttt{LEGS} are points the agent located on the person. The tripwire fires when the gripper or the base's edge comes near them, when the VLM sees the person reach toward the gripper, and every 8\,s. It fired on the reach 5\,s after it was set; the agent drew the arm back, located the plate the person had slid, and placed the lemon wedge on it.}
\label{fig:supp_example}
\end{figure}

\Cref{fig:supp_example} shows a tripwire from an object-state scene in simulation in which the person slides the plate the robot is about to place a lemon wedge on.
The policy had driven the base toward the person twice; the agent moved the base back by 0.1\,m and set this tripwire for the rest of the placement.

\begin{figure}[t]
\fontsize{5.6}{6.5}\selectfont
\begin{verbatim}
def monitor(state, memory):
    n, seen, uf, uw, low = (memory if memory is not None
                            else (0, None, 0, 0, 800.))
    q = ("A white robot arm with a black gripper works at a wooden "
         "table with a blue mug and a black mug tree. Is a person's "
         "hand or arm close to the robot arm or its gripper -- "
         "within about a hand's length of touching it? A hand in "
         "the picture but well away from the robot does not count. "
         "Reply with exactly one line: ANSWER: yes or ANSWER: no")
    r = re.search(r'ANSWER:\s*(yes|no)',
                  vlm(q,[state['images']['front']],8), re.I)
    front = None if r is None else r.group(1).lower() == 'yes'
    uf = uf+1 if front is None else 0
    wrist = False
    if n % 3 == 0:
        q = ("The view from a camera on a robot's gripper, looking "
             "down past its two black fingers at a wooden table. Is "
             "any part of a person visible -- a hand, fingers, an "
             "arm, a sleeve? Reply with exactly one line: "
             "ANSWER: yes or ANSWER: no")
        r = re.search(r'ANSWER:\s*(yes|no)',
                      vlm(q,[state['images']['wrist']],8), re.I)
        wrist = None if r is None else r.group(1).lower() == 'yes'
        uw = uw+1 if wrist is None else 0
    near = bool(front) or bool(wrist) or max(uf,uw)>=2
    if near: seen = state['t']
    grip = (state['gripper'] if state['gripper'] is not None
            else 800.)
    low = min(low,grip)
    mem = (n+1,seen,uf,uw,low)
    if near:
        return mem, ('person near arm in front' if front
                     else 'person at gripper in wrist' if wrist
                     else 'unreadable safety answers')
    if low < 200 and grip > 350:
        return mem, ('gripper reopened after closing: '
                     'check mug hanging')
    if state['t_since_turn'] >= 8:
        return mem, '8 seconds: inspect progress'
    return mem, False
\end{verbatim}
\caption{The tripwire the agent set at its first turn of the mug episode of \cref{fig:real_episode}, with line breaks added. It asks the VLM whether a hand is close to the arm in the front picture at every check and whether a person is visible in the wrist picture at every third check; it also fires when the gripper reopens after closing, and after 8\,s.}
\label{fig:supp_real_example}
\end{figure}

\Cref{fig:supp_real_example} shows the first tripwire of the real-robot episode of \cref{fig:real_episode}.
On the real robot the tripwires see the two cameras' pictures at $320\times240$ pixels and ask the VLM yes/no questions answered in at most 8 tokens; the agent sees the pictures at $640\times480$ pixels.

\section{Additional results}
\label{sec:more}

\paragraph{Contacts over all scenes.}
Over all 168 scenes, without the restriction of the safety metric to the scenes solved without the person, XR touches the person in 86 episodes alone, in 66 with the agentic harness baseline and in 32 with TRIP.
TRIP has the fewest contacts with every policy and in every disturbance type.

\paragraph{How episodes end.}
An episode of TRIP that avoids contact does not always complete the task: with XR, 71 episodes with the person end in neither success nor contact (42\%), against 45 without the person (27\%).
Of these 71, the agent ends 64 by calling \texttt{finish} (41 as stuck, 23 as failure) and 7 end at the spend cap.
No TRIP episode reaches 100 tool calls; the most is 60.
For the baseline the 50 such episodes end with \texttt{finish} in 44 cases and at a limit in 6.

\input{tables/supp_claude}

\paragraph{Claude Opus 5.5 as the agent.}
We ran TRIP with XR and the person with Claude Opus 5.5 as the agent, at its default reasoning effort (medium), with a budget of \$0.80 per episode, on 40 scenes chosen at random among the scenes of \cref{fig:safety}, ten per disturbance type (\cref{tab:supp_claude}).
It succeeds in 13 of the 40 scenes, against 15 with GPT-6.1-sol ($p=0.75$, paired), and touches the person in 17, against 9 with GPT-6.1-sol ($p=0.021$) and 13 for the agentic harness baseline with GPT-6.1-sol.
Claude Opus 5.5 writes 2.2 tripwires per episode against 4.1 for GPT-6.1-sol and makes half as many tool calls per minute; its tool calls take 5.0\,s of agent time on average, against 8.1\,s.
Of its 17 contacts, three come within the first second of the episode, all in task-state scenes.

\section{RoboHD details}
\label{sec:robohd_detail}

\input{tables/supp_tasks}

\begin{figure*}[t]
\centering
\includegraphics[width=\textwidth]{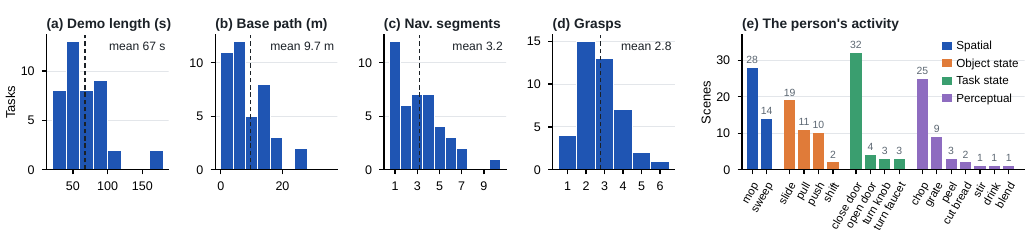}
\caption{RoboHD in numbers. (a)--(d) The source demonstration of each of the 42 tasks: its length, the length of the base's path, and the number of navigation segments and of grasps. (e) The person's activity in the 168 scenes, by disturbance type.}
\label{fig:supp_robohd_stats}
\end{figure*}

\paragraph{Tasks.}
\Cref{tab:supp_tasks} lists the 42 tasks and the person's activity in each scene, and \cref{fig:supp_robohd_stats} summarizes them.
They are composite tasks of RoboCasa365 in which the robot navigates between fixtures; we applied no other rule.

\input{tables/supp_timing}

\paragraph{Scenes.}
Each scene takes its kitchen, objects and instruction from the task, and the avatar's activity from a clip placed in that kitchen.
A scene is kept only if its disturbance has the timing of \cref{tab:supp_timing} when the task's source demonstration is replayed with the avatar's controller running.
At the start of an episode the avatar's stand point and the thresholds of its triggers are drawn once from the scene's seed (the stand point within 0.3\,m and 20$^\circ$ in spatial scenes and within 0.15\,m and 10$^\circ$ in perceptual scenes).
In perceptual scenes the avatar starts at least 0.25\,m from the robot.
No scene has a dead end: the robot can always move out of the person's way.

\paragraph{The avatar's controller.}
The avatar's stages advance on the robot's progress (\cref{sec:dataset}), so a robot that never reaches a step of the task does not meet that step's disturbance.
In all 42 spatial scenes and in 37 of the 42 perceptual scenes the avatar is already at its activity when the episode starts; in object-state and task-state scenes it mostly waits beside the object or fixture, or walks up when the robot nears it.
In object-state and task-state scenes the avatar's hand moves the object or the fixture's joint directly along the clip's motion, since the avatar has no collision body; the robot's policy and the task's check then see the new state.

%% file: tables/supp_tools.tex
\begin{table}[t]
\centering
\footnotesize
\setlength{\tabcolsep}{3pt}
\begin{tabular}{@{}lp{0.70\linewidth}@{}}
\toprule
Call & What it does \\
\midrule
\multicolumn{2}{@{}l}{\emph{Observation}} \\
\texttt{view} & Pictures of the named cameras; no robot time passes. \\
\texttt{locate} & 3D points of up to 10 pixels of one camera, from its depth, in the robot's frame and in the kitchen's fixed frame. \\
\midrule
\multicolumn{2}{@{}l}{\emph{Policy execution}} \\
\texttt{vla} & Run the policy with the task instruction. TRIP: until a tripwire fires. Baseline: for 1--16\,s. \\
\midrule
\multicolumn{2}{@{}l}{\emph{Direct action}} \\
\texttt{move\_base} & Drive forward and sideways (each within 1\,m) and turn (within 90$^\circ$). \\
\texttt{move\_arm\_to} & Move the gripper to a point; open, close or keep the grip. \\
\texttt{move\_arm\_by} & Move the gripper by an offset (each axis within 0.3\,m). \\
\texttt{gripper} & Open or close the gripper. \\
\texttt{grasp\_at} & Open, descend from 10\,cm above a point, close, lift 15\,cm. \\
\texttt{wait} & Hold still. TRIP: until a tripwire fires. Baseline: for 0.4--3\,s. \\
\midrule
\multicolumn{2}{@{}l}{\emph{Tripwire programming} (TRIP only)} \\
\texttt{monitor} & Replace the tripwire with new code on the named cameras; empty code restores the default. \\
\midrule
\texttt{finish} & End the episode as success, failure or stuck. \\
\bottomrule
\end{tabular}
\caption{The agent's tool calls, by the tool groups of \cref{sec:agent}. The agentic harness baseline has every call except \texttt{monitor}.}
\label{tab:supp_tools}
\end{table}

%% file: tables/supp_claude.tex
\begin{table}[!tb]
\centering
\footnotesize
\setlength{\tabcolsep}{3pt}
\begin{tabular}{@{}llccccc@{}}
\toprule
Method & Agent & S & C & Tripw./ep & Calls/min & \$/ep \\
\midrule
XR alone & -- & 6 & 24 & -- & -- & -- \\
Agentic & GPT-6.1-sol & 13 & 13 & -- & 30.7 & 0.17 \\
\midrule
\multirow{2}{*}{TRIP} & GPT-6.1-sol & 15 & 9 & 4.1 & 20.6 & 0.23 \\
 & Claude Opus 5.5 & 13 & 17 & 2.2 & 10.5 & 0.35 \\
\bottomrule
\end{tabular}
\caption{XR with the person on 40 scenes, ten per disturbance type. S and C are successes and episodes with contact; tripwires written per episode, tool calls per minute of robot time and API cost per episode.}
\label{tab:supp_claude}
\end{table}

%% file: tables/supp_tasks.tex
\begin{table*}[t]
\centering
\scriptsize
\setlength{\tabcolsep}{2.2pt}
\renewcommand{\arraystretch}{0.95}
\resizebox{\textwidth}{!}{%
\begin{tabular}{@{}lllll@{\hspace{10pt}}lllll@{}}
\toprule
Task & Spatial & Object state & Task state & Perceptual & Task & Spatial & Object state & Task state & Perceptual \\
\midrule
AddLemonToFish & mops & slides plate & closes fridge & stirs & PackFruitContainer & sweeps & slides container & closes fridge & grates \\
ArrangeBreadBasket$^\ast$ & sweeps & slides basket & closes cabinet & chops & PackIdenticalLunches$^\ast$ & mops & shifts container & closes fridge & grates \\
ArrangeBreadBowl & sweeps & slides bowl & closes toaster oven & chops & PlaceMeatInMarinade & mops & slides pan & opens fridge & chops \\
ArrangeTeaAccompaniments & sweeps & slides tray & closes fridge & grates & PlateStoreDinner & mops & pulls plate & closes fridge & chops \\
AssembleCookingArray & mops & slides pan & closes cabinet & peels & PrepareSausageCheese & mops & slides cutting board & opens fridge & grates \\
BuildAppetizerPlate & mops & slides plate & closes fridge & chops & PrepareSmoothie & mops & pushes plate & closes fridge & blends \\
ClearFreezer & mops & pulls bowl & closes freezer & peels & PrewashFoodAssembly & sweeps & pushes bowl & closes cabinet & grates \\
ClearReceptaclesForCleaning & mops & slides glass & turns faucet on & cuts bread & PrewashFoodSorting & sweeps & slides bowl & closes cabinet & chops \\
DeliverStraw$^\ast$ & mops & slides cup & closes drawer & chops & RetrieveMeat & mops & pulls cutting board & closes fridge & chops \\
DisplayMeatVariety & mops & pushes tray & closes fridge & drinks & SanitizePrepCuttingBoard & mops & slides brush & turns faucet on & chops \\
DivideBuffetTrays & sweeps & slides tray & closes fridge & chops & ServeTea & mops & pulls saucer & turns burner knob & chops \\
GarnishCupcake & mops & pushes plate & opens dishwasher & chops & SetBowlsForSoup & mops & pulls plate & closes cabinet & chops \\
GatherTableware$^\ast$ & sweeps & slides glass & closes cabinet & chops & SetUpCuttingStation$^\ast$ & mops & pulls cutting board & closes drawer & chops \\
GetToastedBread$^\ast$ & mops & pulls plate & turns toaster knob & chops & SetupButterPlate & mops & pushes plate & closes fridge & chops \\
HotDogSetup & mops & pulls plate & closes cabinet & chops & SetupSodaBowl & mops & pushes bowl & closes fridge & chops \\
LoadCondimentsInFridge & sweeps & slides ketchup bottle & closes fridge & chops & SortingCleanup & sweeps & pushes bowl & closes cabinet & grates \\
MakeIceLemonade$^\ast$ & mops & pushes glass & opens fridge & grates & SteamInMicrowave$^\ast$ & sweeps & pushes bowl & closes microwave & grates \\
MakeLoadedPotato & mops & slides cutting board & closes microwave & chops & StoreDumplings & mops & pulls container & closes fridge & chops \\
MixedFruitPlatter & sweeps & pushes plate & closes fridge & grates & StoreLeftoversInBowl$^\ast$ & mops & pulls bowl & closes fridge & chops \\
MultistepSteaming & sweeps & shifts pot & turns burner knob & cuts bread & SweetenCoffee & mops & slides mug & closes fridge & peels \\
PackFoodByTemp & mops & slides container & closes fridge & chops & WashFish & sweeps & pulls tray & turns faucet off & chops \\
\bottomrule
\end{tabular}}
\caption{The 42 RoboCasa365 tasks of RoboHD and what the person does in each of the four scenes of a task: mops or sweeps the robot\textquotesingle s route (spatial), moves an object the task needs (object state), changes a fixture\textquotesingle s state (task state), or does an activity of their own beside the robot (perceptual). $^\ast$Task of the target split.}
\label{tab:supp_tasks}
\end{table*}

%% file: tables/supp_timing.tex
\begin{table}[!tb]
\centering
\footnotesize
\setlength{\tabcolsep}{3pt}
\begin{tabular}{@{}lp{0.74\linewidth}@{}}
\toprule
Type & The disturbance counts when \\
\midrule
Spatial & the person is on the robot's route for at least 3\,s before the base comes close. \\
Object state & the person's push ends while the robot still holds the object it is about to place (36 scenes), or before the gripper reaches the object the person moved (6). \\
Task state & the person's change of the fixture ends at least 1.5\,s before the gripper reaches it (39 scenes), or while the robot uses it (3). \\
Perceptual & the person is within 1\,m of the robot and in one of its front views for at least 3\,s while it manipulates. \\
\bottomrule
\end{tabular}
\caption{The timing each scene's disturbance must meet in the source demonstration.}
\label{tab:supp_timing}
\end{table}

%% file: main.bib
@article{brohan2022rt,
  title={Rt-1: Robotics transformer for real-world control at scale},
  author={Brohan, Anthony and Brown, Noah and Carbajal, Justice and Chebotar, Yevgen and Dabis, Joseph and Finn, Chelsea and Gopalakrishnan, Keerthana and Hausman, Karol and Herzog, Alex and Hsu, Jasmine and others},
  journal={arXiv preprint arXiv:2212.06817},
  year={2022}
}

@article{brohan2023rt,
  title={Rt-2: Vision-language-action models transfer web knowledge to robotic control},
  author={Brohan, Anthony and Brown, Noah and Carbajal, Justice and Chebotar, Yevgen and Chen, Xi and Choromanski, Krzysztof and Ding, Tianli and Driess, Danny and Dubey, Avinava and Finn, Chelsea and others},
  journal={arXiv preprint arXiv:2307.15818},
  year={2023}
}

@inproceedings{openx2024,
  title={Open x-embodiment: Robotic learning datasets and rt-x models: Open x-embodiment collaboration 0},
  author={O’Neill, Abby and Rehman, Abdul and Maddukuri, Abhiram and Gupta, Abhishek and Padalkar, Abhishek and Lee, Abraham and Pooley, Acorn and Gupta, Agrim and Mandlekar, Ajay and Jain, Ajinkya and others},
  booktitle={2024 IEEE International Conference on Robotics and Automation (ICRA)},
  pages={6892--6903},
  year={2024},
  organization={IEEE}
}

@article{octo2024,
  title={Octo: An open-source generalist robot policy},
  author={Team, Octo Model and Ghosh, Dibya and Walke, Homer and Pertsch, Karl and Black, Kevin and Mees, Oier and Dasari, Sudeep and Hejna, Joey and Kreiman, Tobias and Xu, Charles and others},
  journal={arXiv preprint arXiv:2405.12213},
  year={2024}
}

@article{kim2024openvla,
  title={Openvla: An open-source vision-language-action model},
  author={Kim, Moo Jin and Pertsch, Karl and Karamcheti, Siddharth and Xiao, Ted and Balakrishna, Ashwin and Nair, Suraj and Rafailov, Rafael and Foster, Ethan and Lam, Grace and Sanketi, Pannag and others},
  journal={arXiv preprint arXiv:2406.09246},
  year={2024}
}

@article{bousmalis2023robocat,
  title={Robocat: A self-improving generalist agent for robotic manipulation},
  author={Bousmalis, Konstantinos and Vezzani, Giulia and Rao, Dushyant and Devin, Coline and Lee, Alex X and Bauz{\'a}, Maria and Davchev, Todor and Zhou, Yuxiang and Gupta, Agrim and Raju, Akhil and others},
  journal={arXiv preprint arXiv:2306.11706},
  year={2023}
}

@article{black2024pi_0,
  title={$\pi_0$: A Vision-Language-Action Flow Model for General Robot Control},
  author={Black, Kevin and Brown, Noah and Driess, Danny and Esmail, Adnan and Equi, Michael and Finn, Chelsea and Fusai, Niccolo and Groom, Lachy and Hausman, Karol and Ichter, Brian and others},
  journal={arXiv preprint arXiv:2410.24164},
  year={2024}
}

@article{intelligence2025pi_,
  title={$\pi_{0.5}$: a Vision-Language-Action Model with Open-World Generalization},
  author={Intelligence, Physical and Black, Kevin and Brown, Noah and Darpinian, James and Dhabalia, Karan and Driess, Danny and Esmail, Adnan and Equi, Michael and Finn, Chelsea and Fusai, Niccolo and others},
  journal={arXiv preprint arXiv:2504.16054},
  year={2025}
}

@article{bjorck2025gr00t,
  title={Gr00t n1: An open foundation model for generalist humanoid robots},
  author={Bjorck, Johan and Casta{\~n}eda, Fernando and Cherniadev, Nikita and Da, Xingye and Ding, Runyu and Fan, Linxi and Fang, Yu and Fox, Dieter and Hu, Fengyuan and Huang, Spencer and others},
  journal={arXiv preprint arXiv:2503.14734},
  year={2025}
}

@article{team2026xiaomi,
  title={Xiaomi-Robotics-1: Scaling Vision-Language-Action Models with over 100K Hours of Real-World Trajectories},
  author={Team, Xiaomi Robotics and Guo, Jun and Jin, Piaopiao and Li, Jason and Li, Peiyan and Li, Yingyan and Liu, Futeng and Peng, Wanli and Qin, Optimus and Su, Yifei and others},
  journal={arXiv preprint arXiv:2607.15330},
  year={2026}
}

@article{marrie2026reki,
  title={REKI: Registering Knowledge in VLAs},
  author={Marrie, Juliette and Pacaud, Paul and Rahary, Adrien Ramanana and Chen, Shizhe and B{\"o}hle, Moritz},
  journal={arXiv preprint arXiv:2610.22178},
  year={2026}
}

@article{chi2023diffusion,
  title={Diffusion policy: Visuomotor policy learning via action diffusion},
  author={Chi, Cheng and Xu, Zhenjia and Feng, Siyuan and Cousineau, Eric and Du, Yilun and Burchfiel, Benjamin and Tedrake, Russ and Song, Shuran},
  journal={The International Journal of Robotics Research},
  volume={44},
  number={10-11},
  pages={1684--1704},
  year={2025},
  publisher={Sage Publications, Sage UK: London, England}
}

@inproceedings{liu2024rdt,
  title={Rdt-1b: a diffusion foundation model for bimanual manipulation},
  author={Liu, Songming and Wu, Lingxuan and Li, Bangguo and Tan, Hengkai and Chen, Huayu and Wang, Zhengyi and Xu, Ke and Su, Hang and Zhu, Jun},
  booktitle={International Conference on Learning Representations},
  volume={2025},
  pages={29982--30009},
  year={2025}
}

@inproceedings{jiang2022vima,
  title={{VIMA}: Robot Manipulation with Multimodal Prompts},
  author={Jiang, Yunfan and Gupta, Agrim and Zhang, Zichen and Wang, Guanzhi and Dou, Yongqiang and Chen, Yanjun and Fei-Fei, Li and Anandkumar, Anima and Zhu, Yuke and Fan, Linxi},
  booktitle={International Conference on Machine Learning (ICML)},
  pages={14975--15022},
  year={2023}
}

@article{zhao2023learning,
  title={Learning fine-grained bimanual manipulation with low-cost hardware},
  author={Zhao, Tony Z and Kumar, Vikash and Levine, Sergey and Finn, Chelsea},
  journal={arXiv preprint arXiv:2304.13705},
  year={2023}
}

@article{ye2026world,
  title={World action models are zero-shot policies},
  author={Ye, Seonghyeon and Ge, Yunhao and Zheng, Kaiyuan and Gao, Shenyuan and Yu, Sihyun and Kurian, George and Indupuru, Suneel and Tan, You Liang and Zhu, Chuning and Xiang, Jiannan and others},
  journal={arXiv preprint arXiv:2602.15922},
  year={2026}
}

@article{wang2026action,
  title={An Action Is Worth One Patch: Unified World-Action Modeling with PatchWAM},
  author={Wang, Tianheng and Xie, Zhou and Jia, Heng and Xu, Jianhua and Zhang, Tong and Yu, Kaicheng},
  journal={arXiv preprint arXiv:2609.25961},
  year={2026}
}

@article{driess2023palme,
  title={Palm-e: An embodied multimodal language model},
  author={Driess, Danny and Xia, Fei and Sajjadi, Mehdi SM and Lynch, Corey and Chowdhery, Aakanksha and Ichter, Brian and Wahid, Ayzaan and Tompson, Jonathan and Vuong, Quan and Yu, Tianhe and others},
  journal={arXiv preprint arXiv:2303.03378},
  year={2023}
}

@article{ahn2022can,
  title={Do as i can, not as i say: Grounding language in robotic affordances},
  author={Ahn, Michael and Brohan, Anthony and Brown, Noah and Chebotar, Yevgen and Cortes, Omar and David, Byron and Finn, Chelsea and Fu, Chuyuan and Gopalakrishnan, Keerthana and Hausman, Karol and others},
  journal={arXiv preprint arXiv:2204.01691},
  year={2022}
}

@inproceedings{liang2023code,
  title={Code as policies: Language model programs for embodied control},
  author={Liang, Jacky and Huang, Wenlong and Xia, Fei and Xu, Peng and Hausman, Karol and Ichter, Brian and Florence, Pete and Zeng, Andy},
  booktitle={2023 IEEE International conference on robotics and automation (ICRA)},
  pages={9493--9500},
  year={2023},
  organization={IEEE}
}

@article{singh2023progprompt,
  title={Progprompt: Generating situated robot task plans using large language models},
  author={Singh, Ishika and Blukis, Valts and Mousavian, Arsalan and Goyal, Ankit and Xu, Danfei and Tremblay, Jonathan and Fox, Dieter and Thomason, Jesse and Garg, Animesh},
  journal={arXiv preprint arXiv:2209.11302},
  year={2022}
}

@article{huang2022inner,
  title={Inner monologue: Embodied reasoning through planning with language models},
  author={Huang, Wenlong and Xia, Fei and Xiao, Ted and Chan, Harris and Liang, Jacky and Florence, Pete and Zeng, Andy and Tompson, Jonathan and Mordatch, Igor and Chebotar, Yevgen and others},
  journal={arXiv preprint arXiv:2207.05608},
  year={2022}
}

@article{huang2023voxposer,
  title={Voxposer: Composable 3d value maps for robotic manipulation with language models},
  author={Huang, Wenlong and Wang, Chen and Zhang, Ruohan and Li, Yunzhu and Wu, Jiajun and Fei-Fei, Li},
  journal={arXiv preprint arXiv:2307.05973},
  year={2023}
}

@article{rana2023sayplan,
  title={Sayplan: Grounding large language models using 3d scene graphs for scalable robot task planning},
  author={Rana, Krishan and Haviland, Jesse and Garg, Sourav and Abou-Chakra, Jad and Reid, Ian and Suenderhauf, Niko},
  journal={arXiv preprint arXiv:2307.06135},
  year={2023}
}

@inproceedings{zha2023droc,
  title={Distilling and retrieving generalizable knowledge for robot manipulation via language corrections},
  author={Zha, Lihan and Cui, Yuchen and Lin, Li-Heng and Kwon, Minae and Arenas, Montserrat Gonzalez and Zeng, Andy and Xia, Fei and Sadigh, Dorsa},
  booktitle={2024 IEEE international conference on robotics and automation (ICRA)},
  pages={15172--15179},
  year={2024},
  organization={IEEE}
}

@article{liu2023reflect,
  title={Reflect: Summarizing robot experiences for failure explanation and correction},
  author={Liu, Zeyi and Bahety, Arpit and Song, Shuran},
  journal={arXiv preprint arXiv:2306.15724},
  year={2023}
}

@article{tsui2026faea,
  title={Demonstration-free robotic control via LLM agents},
  author={Tsui, Brian Y and Fang, Alan Y and Hwu, Tiffany J},
  journal={arXiv preprint arXiv:2601.20334},
  year={2026}
}

@article{fu2026cap,
  title={CaP-X: A framework for benchmarking and improving coding agents for robot manipulation},
  author={Fu, Letian and Yu, Justin and El-Refai, Karim and Kou, Ethan and Xue, Haoru and Huang, Huang and Xiao, Wenli and Wang, Guanzhi and Niu, Dantong and Li, Fei-Fei and others},
  journal={arXiv preprint arXiv:2603.22435},
  year={2026}
}

@article{zhang2026harnessvla,
  title={Harness VLA: Steering Frozen VLAs into Reliable Manipulation Primitives via Memory-Guided Agents},
  author={Zhang, Yixian and Zhang, Huanming and Gao, Feng and Li, Xiao and
          Liu, Zhihao and Zhu, Chunyang and Qiu, Jiaxing and Yan, Yuchen and
          Liu, Jiyuan and Tang, Wenhao and Fang, Zhengru and Nie, Yi and
          Wei, Changxu and Wang, Yu and Ding, Wenbo and Yu, Chao},
  journal={arXiv preprint arXiv:2607.08448},
  year={2026},
  url={https://arxiv.org/abs/2607.08448}
}

@article{chen2026showharness,
  title={Show-harness: Just a vlm agent can play robots},
  author={Chen, Yanzhe and Bai, Zechen and Cao, Zhijun and Zeng, Wenzheng and Lin, Kevin Qinghong and Lin, Yiqi and Liang, Guoqiang and Ma, Kevin Yuchen and Huang, Qiming and Shou, Mike Zheng},
  journal={arXiv preprint arXiv:2609.10522},
  year={2026}
}

@misc{wang2026auriga,
  author       = {{StellarEdge Team}},
  title        = {{Auriga Technical Report: A Task-Agnostic Agent Harness for Zero-Shot Embodied Manipulation}},
  year         = {2026},
  howpublished = {Project website},
  url          = {https://auriga.stelledge.com/}
}

@article{team2026systematically,
  title={Systematically exploring the capabilities of GPT-6 Astra as embodied policies},
  author={Team, Galbot and Chen, Xuchuan and Cheng, Xiaoqian and Deng, Yu and Ding, Lihe and Dong, Shaocong and Gao, Xiangjun and Jia, Haozhe and Li, Zekai and Li, Zhoujian and others},
  journal={arXiv preprint arXiv:2609.38537},
  year={2026}
}

@misc{yang2026astranomy,
  title        = {Astranomy: The study of {GPT-6 Astra} as a native robot policy},
  author       = {Yang, Jenai Xuning and Khetan, Ashna and Zhang, Shun},
  year         = {2026},
  month        = {October},
  organization = {Seattle Robotics Lab (SRL), NVIDIA},
  howpublished = {\url{https://research.nvidia.com/labs/srl/projects/astranomy/}},
  note         = {Blog post},
}

@misc{xu2026harness,
  title={{The Harness}: One Agent Harness for Anything that Acts},
  author={Xu, Sihan and Wang, Zilin and Shen, Hui and Xu, Zhengjie and Jin, Weiyang and Niu, Yuwei and Li, Ye and Zhang, Jiechen and Dai, Yining and Luo, Yunhao and Chai, Wenhao and Chen, Xuweiyi and Liu, Ziwei and Yu, Stella},
  year={2026},
  note={Conference on Robot Learning (CoRL), Late Breaking Results},
  url={https://web.eecs.umich.edu/~stellayu/publication/doc/2026harnessCoRLL.pdf}
}

@inproceedings{guo2024doremi,
  title={Doremi: Grounding language model by detecting and recovering from plan-execution misalignment},
  author={Guo, Yanjiang and Wang, Yen-Jen and Zha, Lihan and Chen, Jianyu},
  booktitle={2024 IEEE/RSJ International Conference on Intelligent Robots and Systems (IROS)},
  pages={12124--12131},
  year={2024},
  organization={IEEE}
}

@inproceedings{zhou2025code,
  title={Code-as-monitor: Constraint-aware visual programming for reactive and proactive robotic failure detection},
  author={Zhou, Enshen and Su, Qi and Chi, Cheng and Zhang, Zhizheng and Wang, Zhongyuan and Huang, Tiejun and Sheng, Lu and Wang, He},
  booktitle={2025 IEEE/CVF Conference on Computer Vision and Pattern Recognition (CVPR)},
  pages={6919--6929},
  year={2025},
  organization={IEEE}
}

@article{xu2025can,
  title={Can we detect failures without failure data? uncertainty-aware runtime failure detection for imitation learning policies},
  author={Xu, Chen and Nguyen, Tony Khuong and Dixon, Emma and Rodriguez, Christopher and Miller, Patrick and Lee, Robert and Shah, Paarth and Ambrus, Rares and Nishimura, Haruki and Itkina, Masha},
  journal={arXiv preprint arXiv:2503.08558},
  year={2025}
}

@article{chen2026volo,
  title={VoLo: A Physical Orchestrator for Open-Vocabulary Long-Horizon Manipulation},
  author={Chen, Siyi and Hadfield, Hugo and Zook, Alex and Uy, Mikaela Angelina and Song, Chan Hee and Coumans, Erwin and Yang, Xuning and Ladhak, Faisal and Qu, Qing and Birchfield, Stan and others},
  journal={arXiv preprint arXiv:2606.07723},
  year={2026}
}

@article{tabuada2007event,
  title={Event-triggered real-time scheduling of stabilizing control tasks},
  author={Tabuada, Paulo},
  journal={IEEE Transactions on Automatic control},
  volume={52},
  number={9},
  pages={1680--1685},
  year={2007},
  publisher={IEEE}
}

@inproceedings{heemels2012introduction,
  title={An introduction to event-triggered and self-triggered control},
  author={Heemels, Wilhelmus PMH and Johansson, Karl Henrik and Tabuada, Paulo},
  booktitle={2012 ieee 51st ieee conference on decision and control (cdc)},
  pages={3270--3285},
  year={2012},
  organization={IEEE}
}

@article{james2020rlbench,
  title={Rlbench: The robot learning benchmark \& learning environment},
  author={James, Stephen and Ma, Zicong and Arrojo, David Rovick and Davison, Andrew J},
  journal={IEEE Robotics and Automation Letters},
  volume={5},
  number={2},
  pages={3019--3026},
  year={2020},
  publisher={IEEE}
}

@article{mees2022calvin,
    author = {Oier Mees and Lukas Hermann and Erick Rosete-Beas and Wolfram Burgard},
    title = {CALVIN: A Benchmark for Language-Conditioned Policy Learning for Long-Horizon Robot Manipulation Tasks},
    journal={IEEE Robotics and Automation Letters (RA-L)},
    volume={7},
    number={3},
    pages={7327-7334},
    year={2022}
}

@inproceedings{gu2023maniskill2,
  title={ManiSkill2: A Unified Benchmark for Generalizable Manipulation Skills},
  author={Gu, Jiayuan and Xiang, Fanbo and Li, Xuanlin and Ling, Zhan and Liu, Xiqiang and Mu, Tongzhou and Tang, Yihe and Tao, Stone and Wei, Xinyue and Yao, Yunchao and Yuan, Xiaodi and Xie, Pengwei and Huang, Zhiao and Chen, Rui and Su, Hao},
  booktitle={International Conference on Learning Representations},
  year={2023}
}

@inproceedings{mandlekar2021robomimic,
  title={What Matters in Learning from Offline Human Demonstrations for Robot Manipulation},
  author={Ajay Mandlekar and Danfei Xu and Josiah Wong and Soroush Nasiriany and Chen Wang and Rohun Kulkarni and Li Fei-Fei and Silvio Savarese and Yuke Zhu and Roberto Mart\'{i}n-Mart\'{i}n},
  booktitle={Conference on Robot Learning (CoRL)},
  year={2021}
}

@article{pumacay2024colosseum,
  title={The colosseum: A benchmark for evaluating generalization for robotic manipulation},
  author={Pumacay, Wilbert and Singh, Ishika and Duan, Jiafei and Krishna, Ranjay and Thomason, Jesse and Fox, Dieter},
  journal={arXiv preprint arXiv:2402.08191},
  year={2024}
}

@inproceedings{fei2026libero,
  title={Libero-plus: A progressive robustness benchmark for visual-language-action models},
  author={Fei, Senyu and Wang, Siyin and Shi, Junhao and Dai, Zihao and Cai, Jikun and Qian, Pengfang and Ji, Li and He, Xinzhe and Zhang, Shiduo and Fei, Zhaoye and others},
  booktitle={Proceedings of the IEEE/CVF Conference on Computer Vision and Pattern Recognition},
  pages={38574--38583},
  year={2026}
}

@article{zhou2025libero,
  title={Libero-pro: Towards robust and fair evaluation of vision-language-action models beyond memorization},
  author={Zhou, Xueyang and Xu, Yangming and Tie, Guiyao and Chen, Yongchao and Zhang, Guowen and Chu, Duanfeng and Zhou, Pan and Sun, Lichao},
  journal={arXiv preprint arXiv:2510.03827},
  year={2025}
}

@article{zhang2026libero,
  title={LIBERO-MAX: Do Robot Policies Adapt When the World Changes?},
  author={Zhang, Yunbei and Jin, Zijian and Liu, Yuanzhe and Wang, Janet and Zhang, Xilun and Zhang, Yuyou and Zhang, Zhenyu and Zhang, Daoan and Niu, Shuaicheng and Li, Gen and others},
  journal={arXiv preprint arXiv:2609.36518},
  year={2026}
}

@inproceedings{fang2026towards,
  title={Towards generalizable robotic manipulation in dynamic environments},
  author={Fang, Heng and Li, Shangru and Wang, Shuhan and Xi, Xuanyang and Liang, Dingkang and Bai, Xiang},
  booktitle={European Conference on Computer Vision},
  pages={314--333},
  year={2026},
  organization={Springer}
}

@inproceedings{shridhar2020alfred,
  title={Alfred: A benchmark for interpreting grounded instructions for everyday tasks},
  author={Shridhar, Mohit and Thomason, Jesse and Gordon, Daniel and Bisk, Yonatan and Han, Winson and Mottaghi, Roozbeh and Zettlemoyer, Luke and Fox, Dieter},
  booktitle={2020 IEEE/CVF conference on computer vision and pattern recognition (CVPR)},
  pages={10737--10746},
  year={2020},
  organization={IEEE}
}

@inproceedings{li2023behavior,
  title={Behavior-1k: A benchmark for embodied ai with 1,000 everyday activities and realistic simulation},
  author={Li, Chengshu and Zhang, Ruohan and Wong, Josiah and Gokmen, Cem and Srivastava, Sanjana and Mart{\'\i}n-Mart{\'\i}n, Roberto and Wang, Chen and Levine, Gabrael and Lingelbach, Michael and Sun, Jiankai and others},
  booktitle={Conference on Robot Learning},
  pages={80--93},
  year={2023},
  organization={PMLR}
}

@inproceedings{padmakumar2022teach,
  title={Teach: Task-driven embodied agents that chat},
  author={Padmakumar, Aishwarya and Thomason, Jesse and Shrivastava, Ayush and Lange, Patrick and Narayan-Chen, Anjali and Gella, Spandana and Piramuthu, Robinson and Tur, Gokhan and Hakkani-Tur, Dilek},
  booktitle={Proceedings of the AAAI Conference on Artificial Intelligence},
  volume={36},
  number={2},
  pages={2017--2025},
  year={2022}
}

@inproceedings{puig2018virtualhome,
  title={Virtualhome: Simulating household activities via programs},
  author={Puig, Xavier and Ra, Kevin and Boben, Marko and Li, Jiaman and Wang, Tingwu and Fidler, Sanja and Torralba, Antonio},
  booktitle={2018 IEEE/cvf conference on computer vision and pattern recognition},
  pages={8494--8502},
  year={2018},
  organization={IEEE}
}

@article{li2022igibson,
  title={igibson 2.0: Object-centric simulation for robot learning of everyday household tasks},
  author={Li, Chengshu and Xia, Fei and Mart{\'\i}n-Mart{\'\i}n, Roberto and Lingelbach, Michael and Srivastava, Sanjana and Shen, Bokui and Vainio, Kent and Gokmen, Cem and Dharan, Gokul and Jain, Tanish and others},
  journal={arXiv preprint arXiv:2108.03272},
  year={2021}
}

@inproceedings{nasiriany2026robocasa365,
  title={RoboCasa365: A Large-Scale Simulation Framework for Training and Benchmarking Generalist Robots},
  author={Soroush Nasiriany and Sepehr Nasiriany and Abhiram Maddukuri and Yuke Zhu},
  booktitle={International Conference on Learning Representations (ICLR)},
  year={2026}
}

@inproceedings{puig2024habitat,
  title={{Habitat 3.0}: A Co-Habitat for Humans, Avatars, and Robots},
  author={Puig, Xavier and Undersander, Eric and Szot, Andrew and Cote, Mikael Dallaire and Yang, Tsung-Yen and Partsey, Ruslan and Desai, Ruta and Clegg, Alexander and Hlavac, Michal and Min, So Yeon and others},
  booktitle={International Conference on Learning Representations},
  year={2024},
  url={https://openreview.net/forum?id=4znwzG92CE}
}

@inproceedings{chang2025partnr,
  author = {Matthew Chang and Gunjan Chhablani and Alexander Clegg and Mikael Dallaire Cote and Ruta Desai and Michal Hlavac and Vladimir Karashchuk and Jacob Krantz and Roozbeh Mottaghi and Priyam Parashar and Siddharth Patki and Ishita Prasad and Xavier Puig and Akshara Rai and Ram Ramrakhya and Daniel Tran and Joanne Truong and John M. Turner and Eric Undersander and Tsung-Yen Yang},
  title = {PARTNR: A Benchmark for Planning and Reasoning in Embodied Multi-agent Tasks},
  booktitle = {International Conference on Learning Representations (ICLR)},
  note = {alphabetical author order},
  year = {2025}
}

@article{zhou2026harp,
  title={HARP-Bench: Benchmarking Robot Manipulation Around Active Humans},
  author={Zhou, Qinhong and Zong, Riley and Lin, Chunru and Hu, Diyi and Bhatti, Shivraj Singh and Zhang, Hongxin and Qiu, Lin and Zeng, Hanqing and Gan, Chuang},
  journal={arXiv preprint arXiv:2611.11123},
  year={2026}
}

@misc{BONES,
  title={{BONES Studio}: Human Motion Capture Datasets},
  author={{BONES Studio}},
  howpublished={\url{https://bones.studio/datasets}},
  year={2026},
  note={Motion capture datasets. Accessed: 2026-11-17},
  url={https://bones.studio/datasets}
}

@misc{SEED,
  title={{BONES-SEED}: Skeletal Everyday Embodied Dataset},
  author={{BONES Studio}},
  howpublished={\url{https://bones.studio/datasets/seed}},
  year={2025},
  note={Human motion capture dataset},
  url={https://bones.studio/datasets/seed}
}

@misc{qwen2026qwen35,
    title  = {{Qwen3.5}: Towards Native Multimodal Agents},
    author = {{Qwen Team}},
    month  = {February},
    year   = {2026},
    url    = {https://qwen.ai/blog?id=qwen3.5}
}

@misc{openai2026astra,
  title={{GPT-6 Astra}: A New Generation of Intelligence},
  author={{OpenAI}},
  year={2026},
  howpublished={\url{https://openai.com/index/gpt-6-astra/}},
  note={Official model announcement}
}
